\documentclass[conference]{IEEEtran}
\IEEEoverridecommandlockouts
\usepackage{cite}
\usepackage[numbers,sort&compress]{natbib}

\usepackage[utf8]{inputenc}
\usepackage{amsmath,amssymb}
\usepackage{graphicx}
\usepackage{bm}
\usepackage{enumerate}
\usepackage{comment}
\usepackage{xcolor}
\usepackage{url}
\usepackage{color,soul}
\usepackage{subcaption}
\usepackage{float}

\definecolor{light-gray}{gray}{0.8}

\usepackage{amsmath,amssymb,amsfonts}
\usepackage{algorithmic}
\usepackage[linesnumbered,ruled]{algorithm2e}
\usepackage{graphicx}
\usepackage{textcomp}
\usepackage{xcolor}
\usepackage{subcaption}

\def\BibTeX{{\rm B\kern-.05em{\sc i\kern-.025em b}\kern-.08em
    T\kern-.1667em\lower.7ex\hbox{E}\kern-.125emX}}

\makeatletter
\newcommand{\linebreakand}{%
  \end{@IEEEauthorhalign}
  \hfill\mbox{}\par
  \mbox{}\hfill\begin{@IEEEauthorhalign}
}
\makeatother

\begin{document}

\title{Data Augmentation Through Random Style Replacement\\}

\author{

\small 

\begin{tabular}[t]{c@{\extracolsep{8em}}c} 

1\textsuperscript{st} Qikai Yang\textsuperscript{*}  & 2\textsuperscript{nd} Cheng Ji \\
\textit{University of Illinois Urbana-Champaign}, Urbana, USA & \textit{University of Illinois Urbana-Champaign}, Urbana, USA \\
\textsuperscript{*}Corresponding Author: qikaiy2@illinois.edu & \\

\\

3\textsuperscript{th} Huaiying Luo & 4\textsuperscript{th} Panfeng Li \\
\textit{Cornell University}, Ithaca, USA & \textit{University of Michigan}, Ann Arbor, USA \\ 

\\

\multicolumn{2}{c}{5\textsuperscript{th} Zhicheng Ding}  \\
\multicolumn{2}{c}{\textit{Columbia University}, New York, USA}\\

\end{tabular}
}

\maketitle

\begin{abstract}
In this paper, we introduce a novel data augmentation technique that combines the advantages of style augmentation and random erasing by selectively replacing image subregions with style-transferred patches. Our approach first applies a random style transfer to training images, then randomly substitutes selected areas of these images with patches derived from the style-transferred versions. This method is able to seamlessly accommodate a wide range of existing style transfer algorithms and can be readily integrated into diverse data augmentation pipelines. By incorporating our strategy, the training process becomes more robust and less prone to overfitting. Comparative experiments demonstrate that, relative to previous style augmentation methods, our technique achieves superior performance and faster convergence.
\end{abstract}


\section{Introduction}
Recent advancements in deep learning have driven significant progress in a wide range of computer vision tasks, including image classification, object detection, and semantic segmentation \cite{dan2024image, li2024segmentation}. Despite these advances, many of these tasks continue to face a fundamental bottleneck: a lack of sufficient labeled data \cite{coates2011analysis, hastie2009unsupervised}. Annotating large-scale datasets is both time-consuming and costly, which can limit the applicability of deep neural networks in specialized or rapidly evolving domains. To mitigate this issue, data augmentation techniques are heavily utilized, artificially expanding and diversifying the training set so that models generalize more effectively.

\begin{figure}[h!]
  \centering
  \begin{subfigure}[T]{0.19\linewidth}
    \includegraphics[width=\linewidth, height=0.6\linewidth]{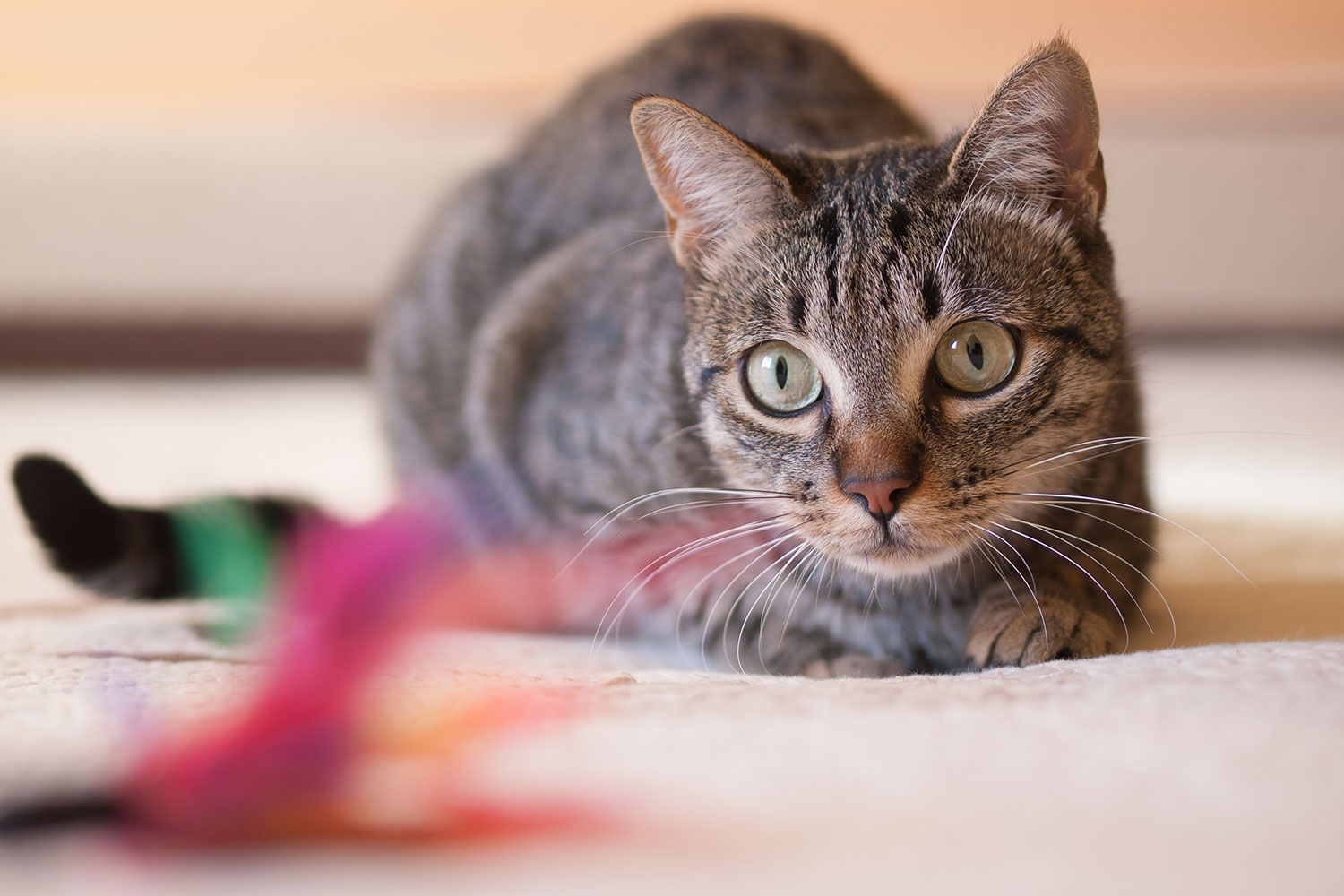}
  \end{subfigure}
  \begin{subfigure}[T]{0.19\linewidth}
    \includegraphics[width=\linewidth, height=0.6\linewidth]{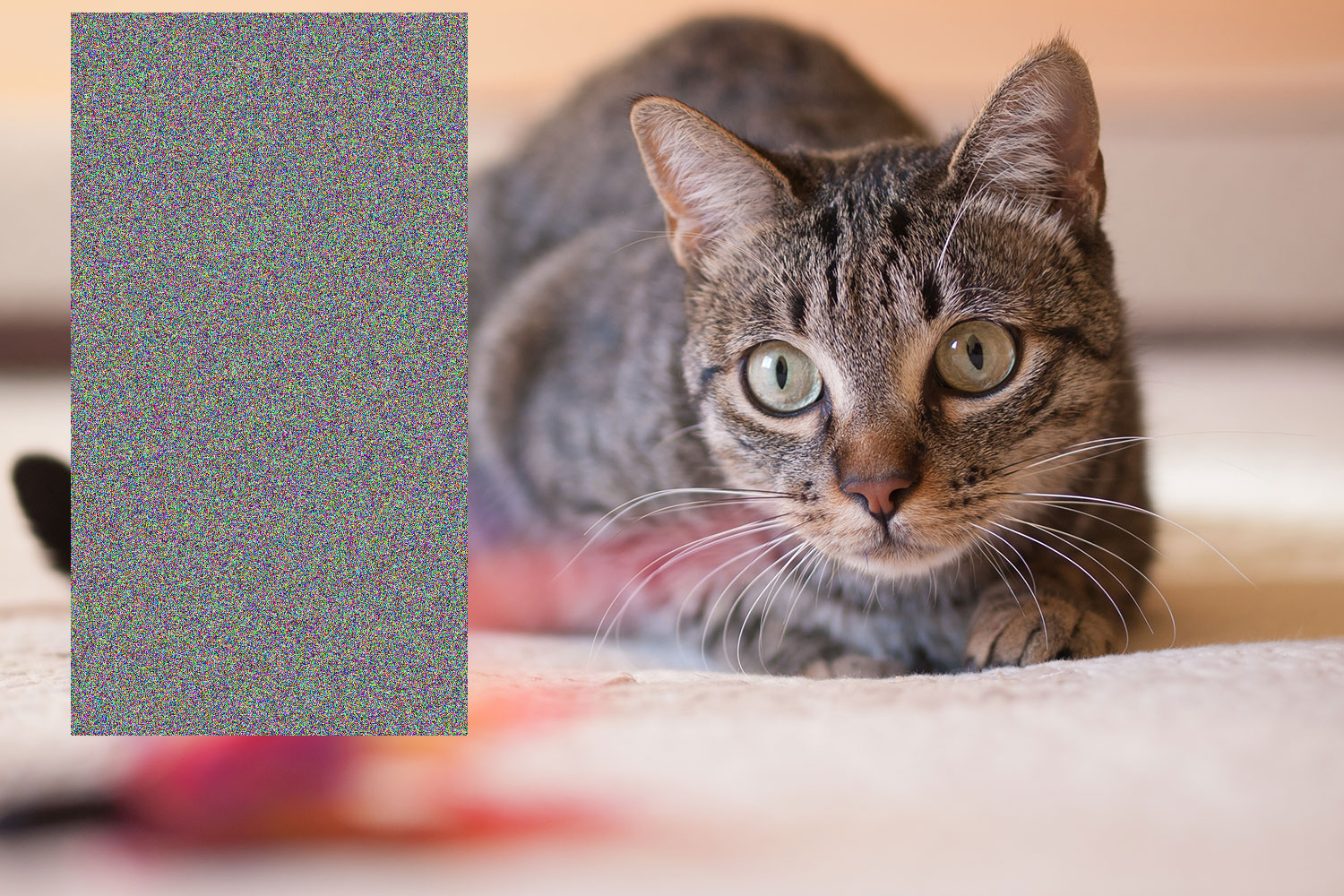}
  \end{subfigure}
  \begin{subfigure}[T]{0.19\linewidth}
    \includegraphics[width=\linewidth, height=0.6\linewidth]{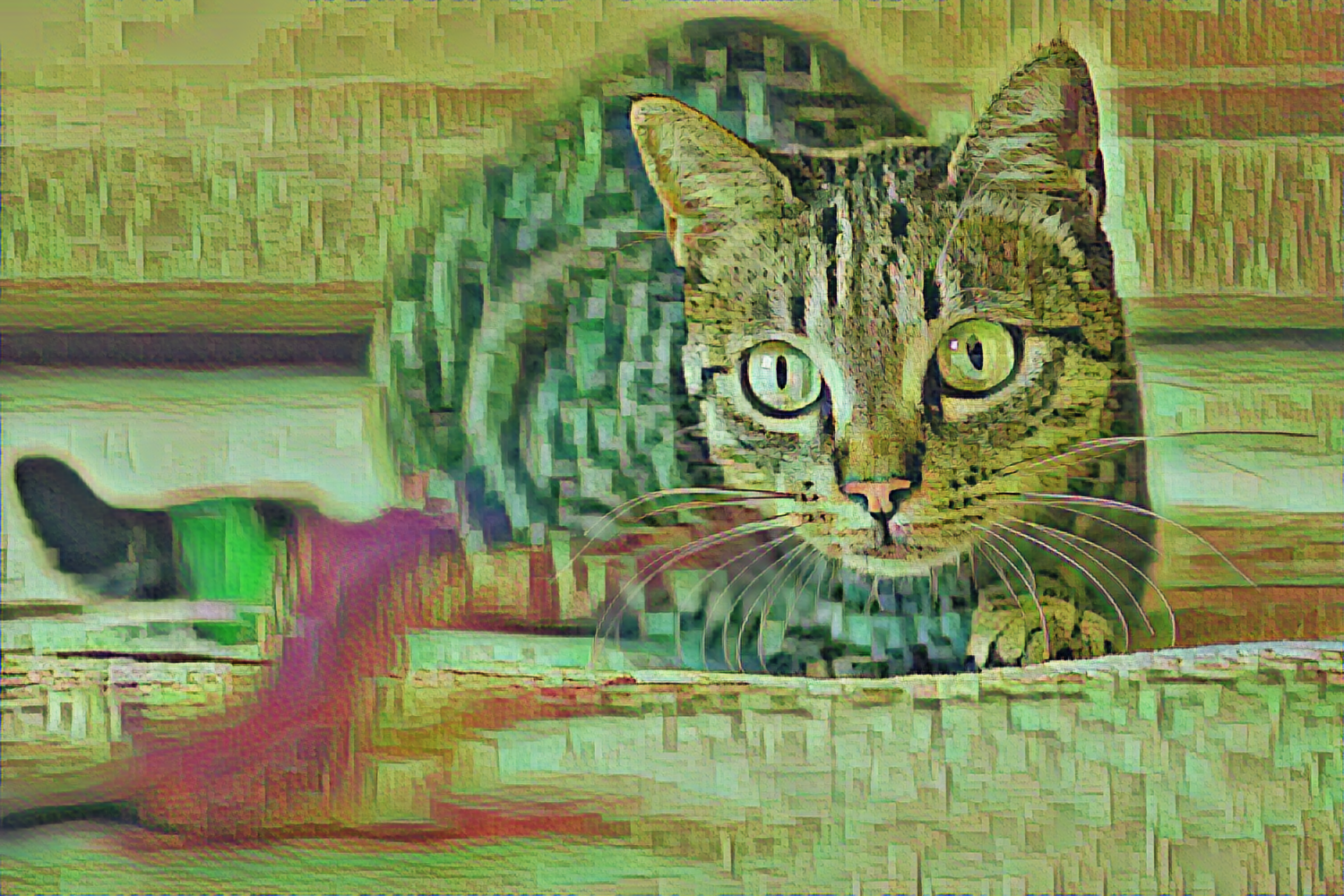}
  \end{subfigure}
  \begin{subfigure}[T]{0.19\linewidth}
    \includegraphics[width=\linewidth, height=0.6\linewidth]{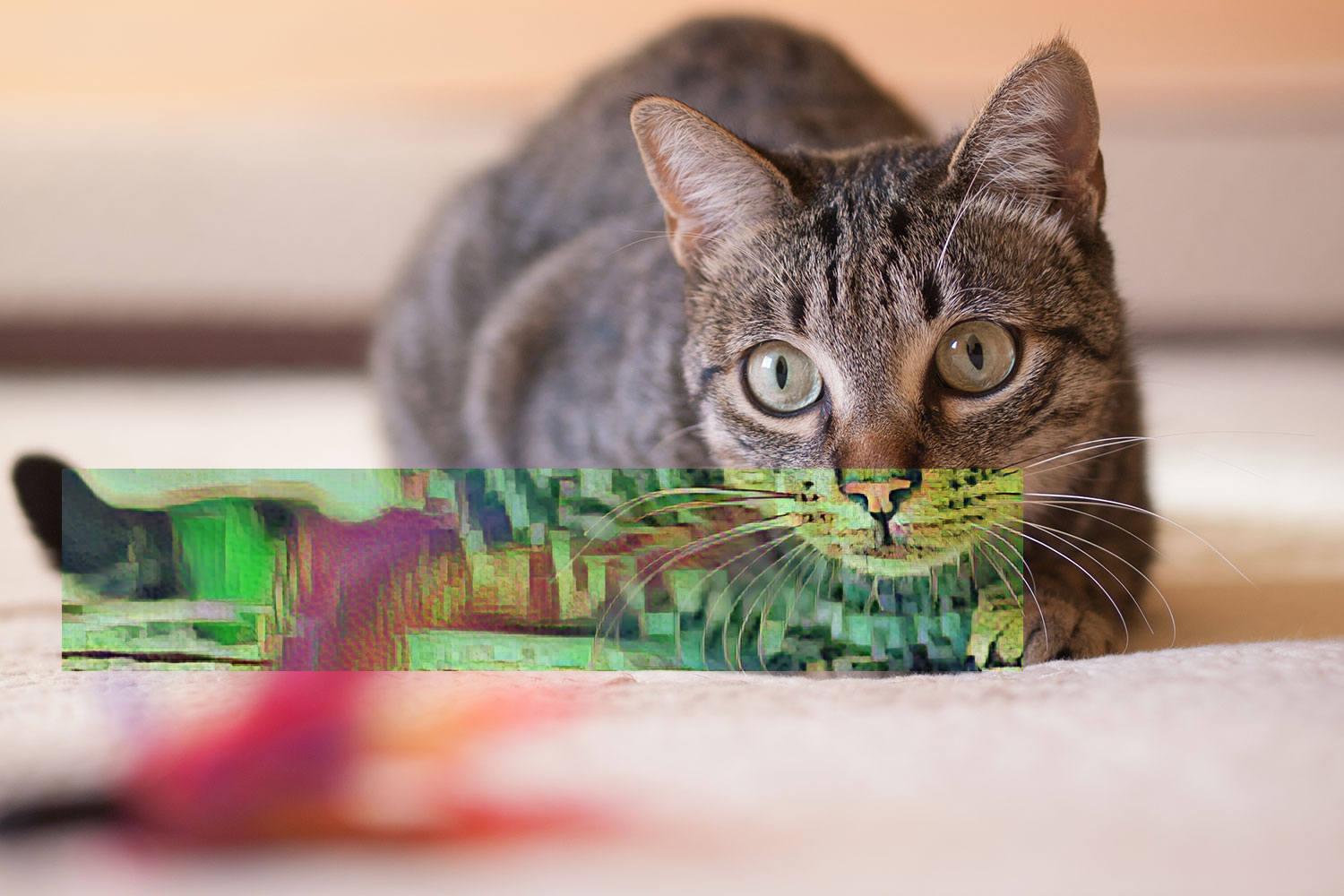}
  \end{subfigure}
  \begin{subfigure}[T]{0.19\linewidth}
    \includegraphics[width=\linewidth, height=0.6\linewidth]{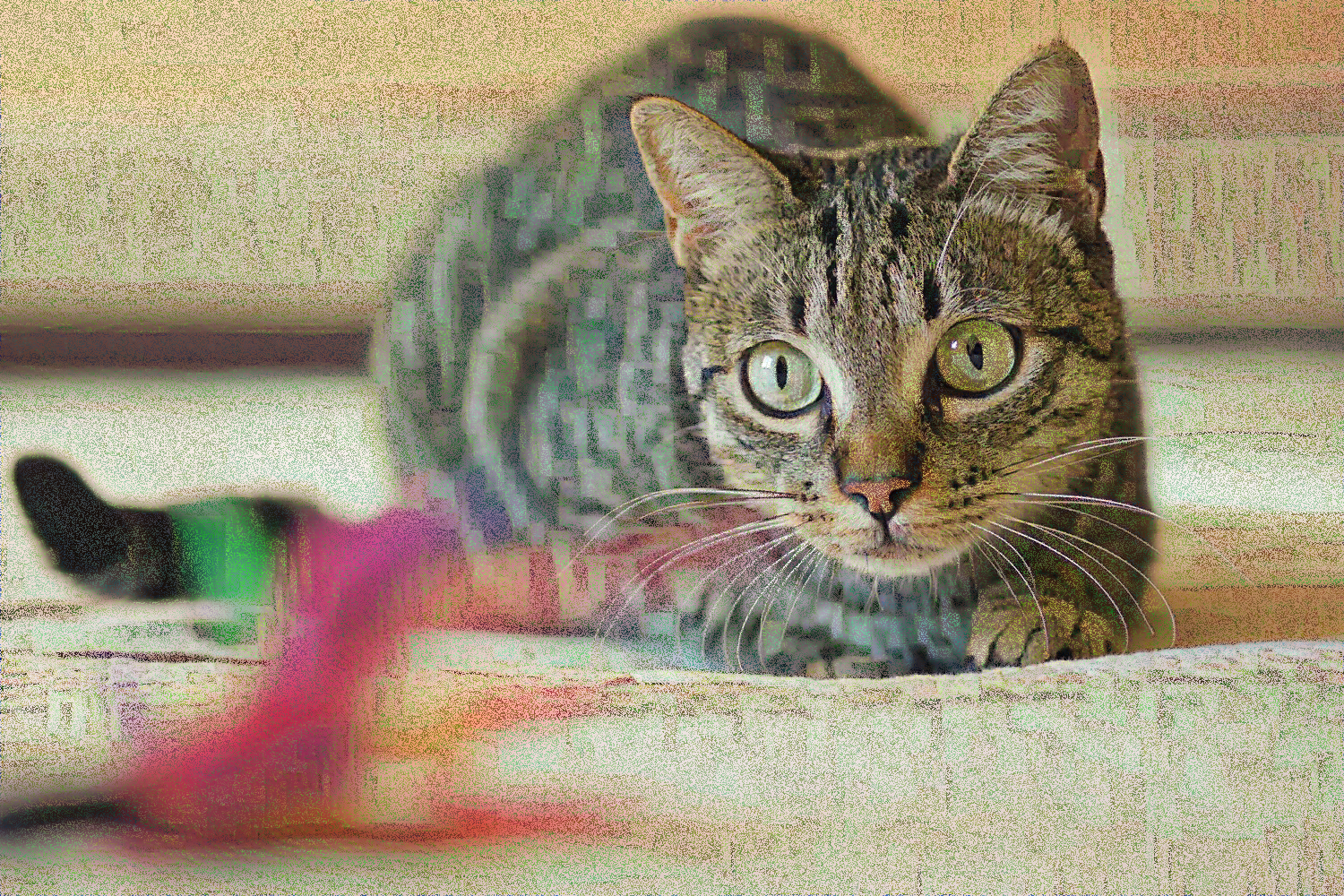}
  \end{subfigure}

  \begin{subfigure}[T]{0.19\linewidth}
    \includegraphics[width=\linewidth, height=0.6\linewidth]{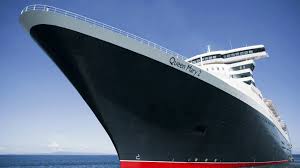}
  \end{subfigure}
  \begin{subfigure}[T]{0.19\linewidth}
    \includegraphics[width=\linewidth, height=0.6\linewidth]{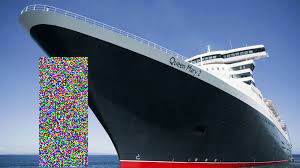}
  \end{subfigure}
  \begin{subfigure}[T]{0.19\linewidth}
    \includegraphics[width=\linewidth, height=0.6\linewidth]{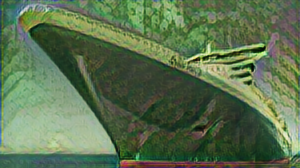}
  \end{subfigure}
  \begin{subfigure}[T]{0.19\linewidth}
    \includegraphics[width=\linewidth, height=0.6\linewidth]{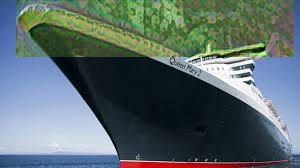}
  \end{subfigure}
  \begin{subfigure}[T]{0.19\linewidth}
    \includegraphics[width=\linewidth, height=0.6\linewidth]{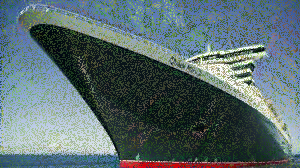}
  \end{subfigure}

  \begin{subfigure}[T]{0.19\linewidth}
    \includegraphics[width=\linewidth, height=0.6\linewidth]{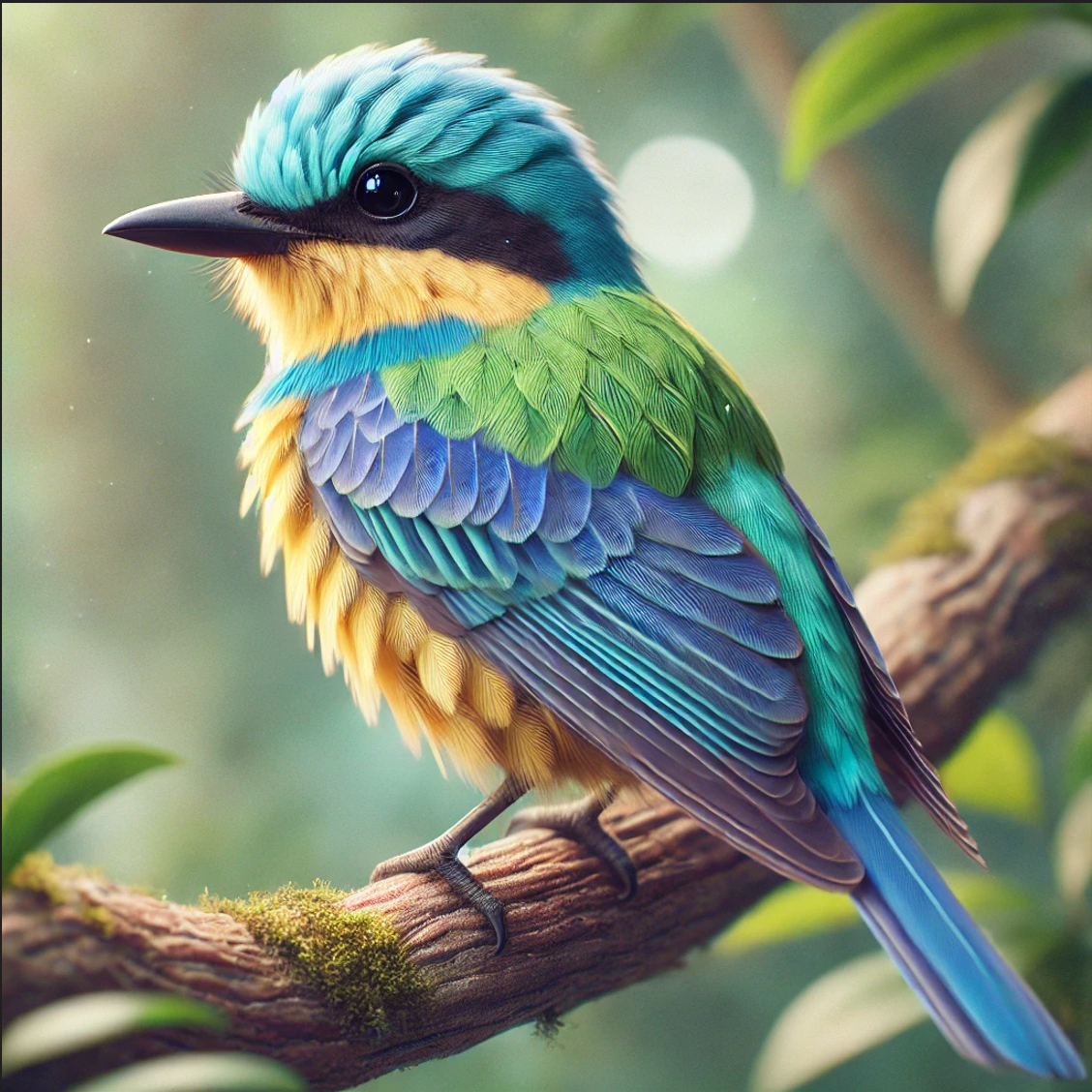}
  \end{subfigure}
  \begin{subfigure}[T]{0.19\linewidth}
    \includegraphics[width=\linewidth, height=0.6\linewidth]{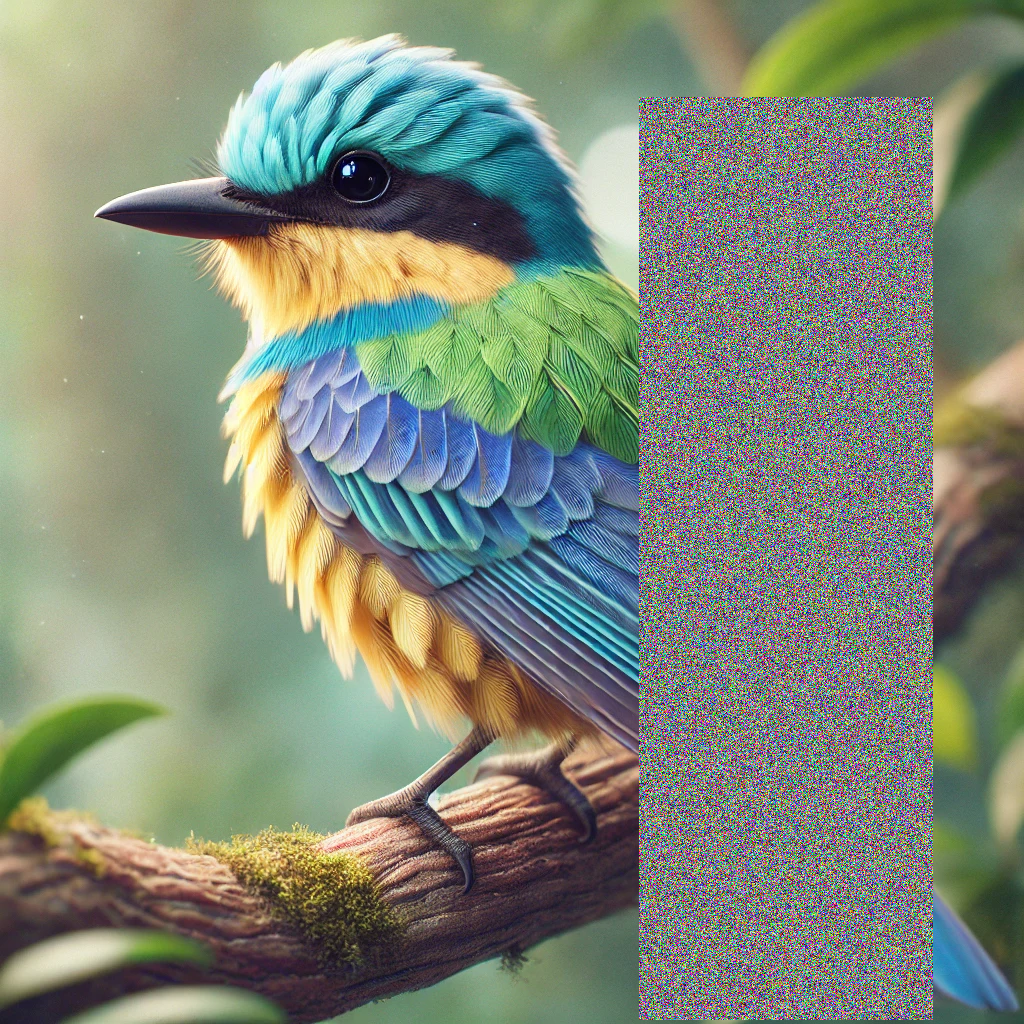}
  \end{subfigure}
  \begin{subfigure}[T]{0.19\linewidth}
    \includegraphics[width=\linewidth, height=0.6\linewidth]{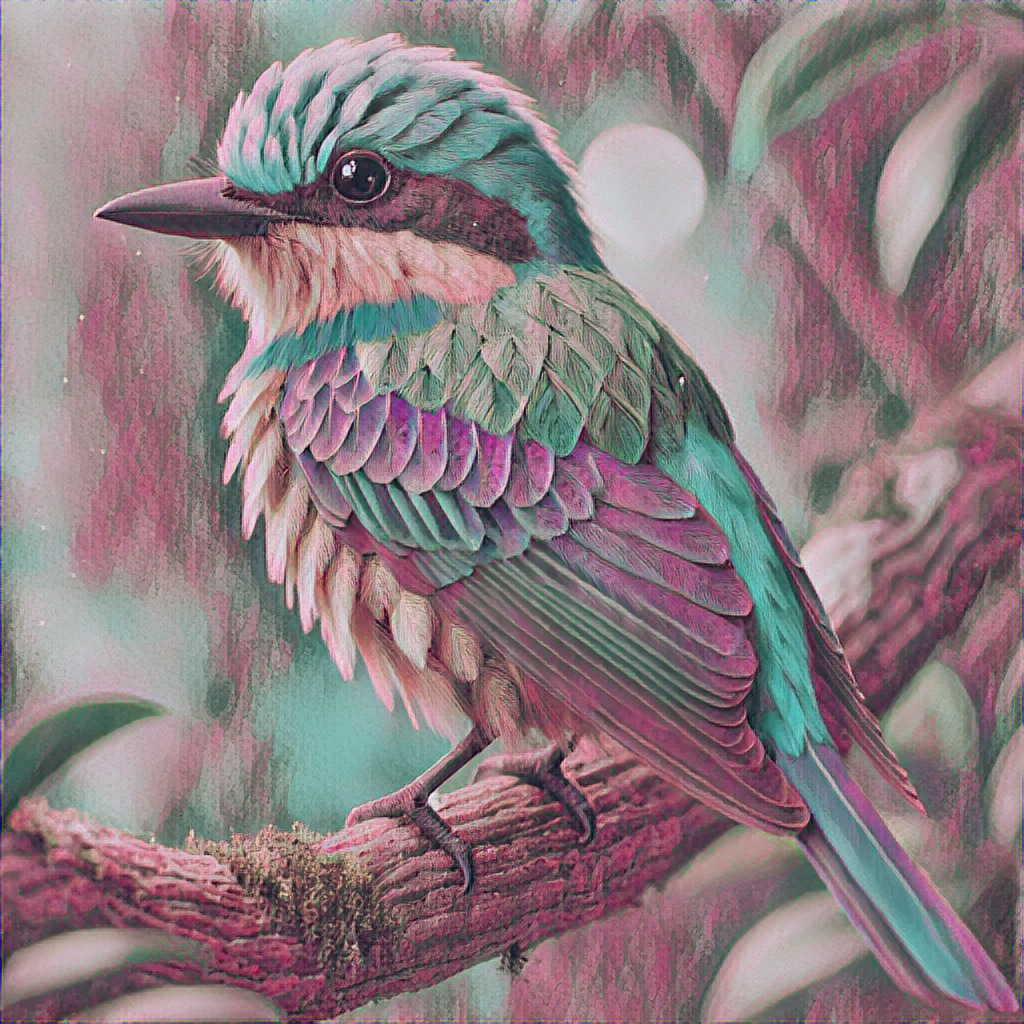}
  \end{subfigure}
  \begin{subfigure}[T]{0.19\linewidth}
    \includegraphics[width=\linewidth, height=0.6\linewidth]{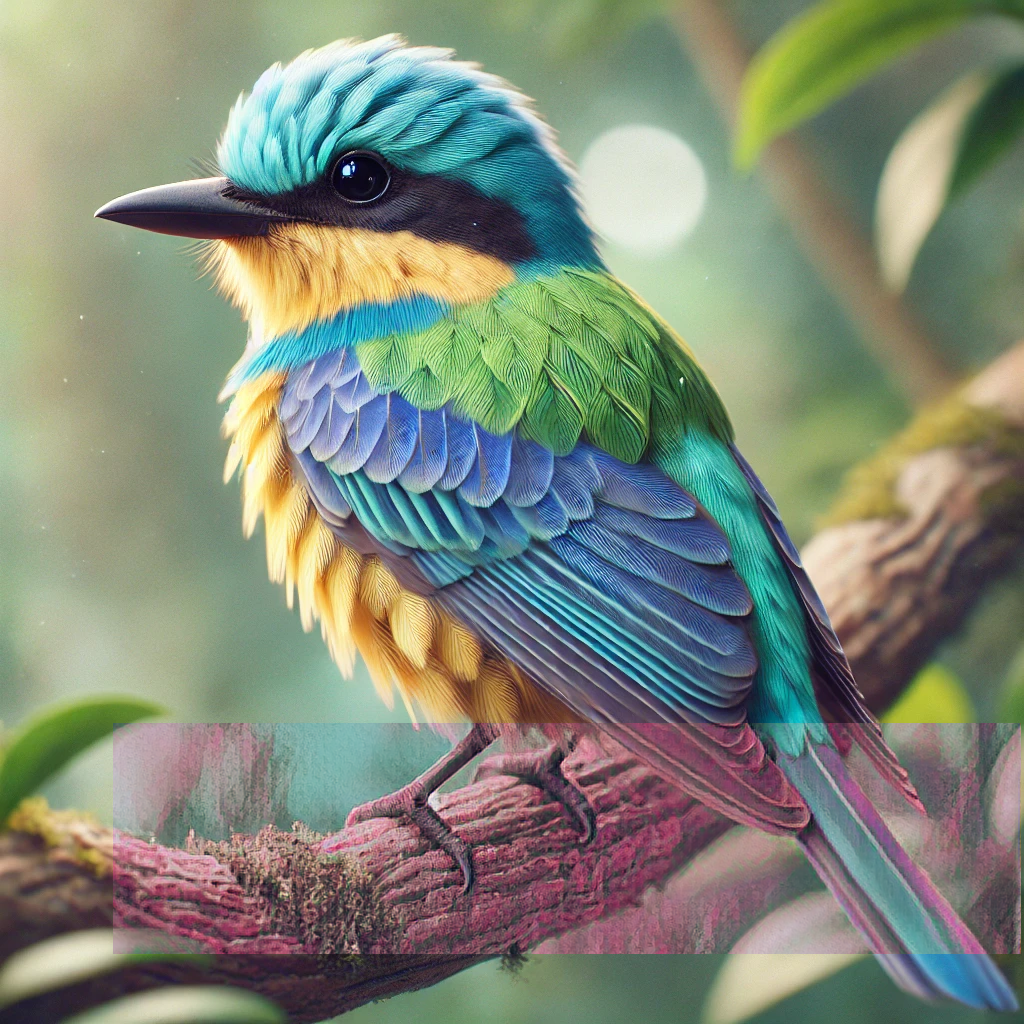}
  \end{subfigure}
  \begin{subfigure}[T]{0.19\linewidth}
    \includegraphics[width=\linewidth, height=0.6\linewidth]{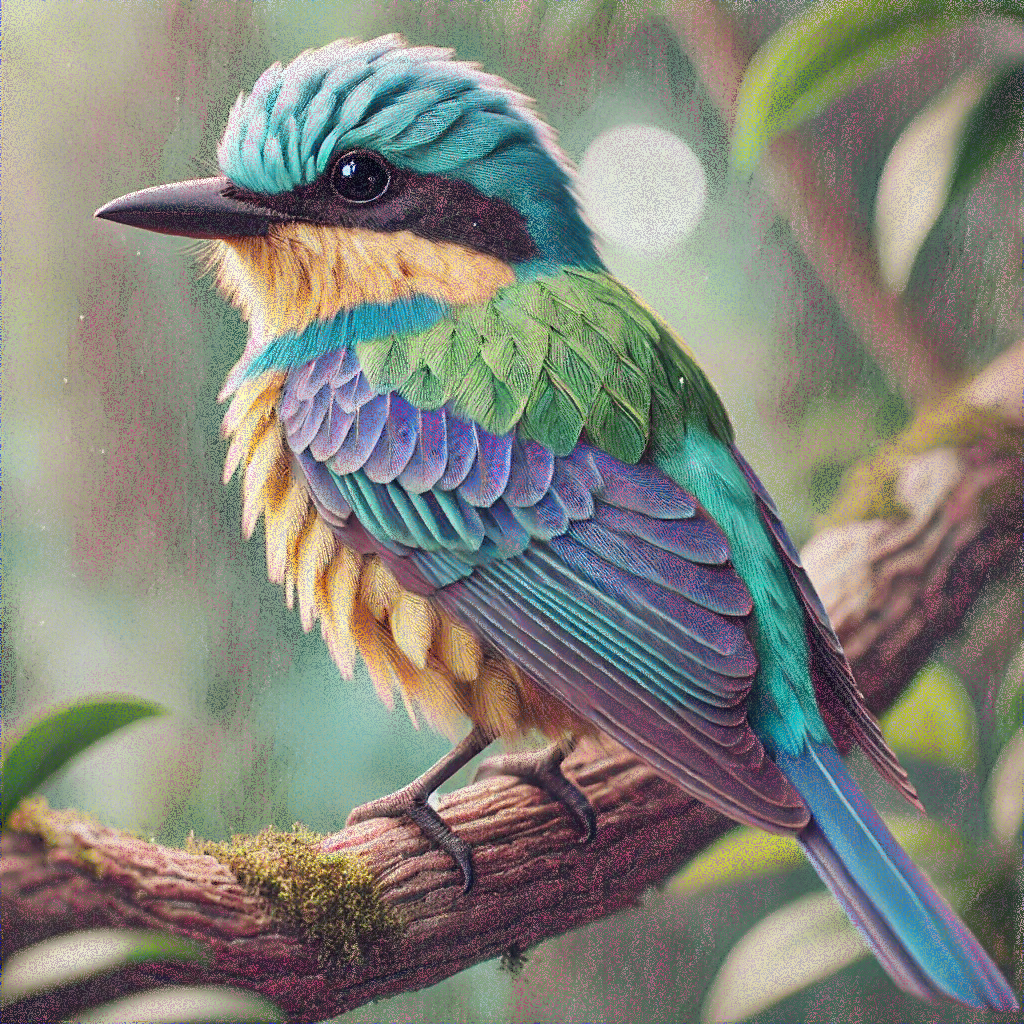}
  \end{subfigure}
  
  \centering
  \begin{subfigure}[T]{0.19\linewidth}
    \includegraphics[width=\linewidth, height=0.6\linewidth]{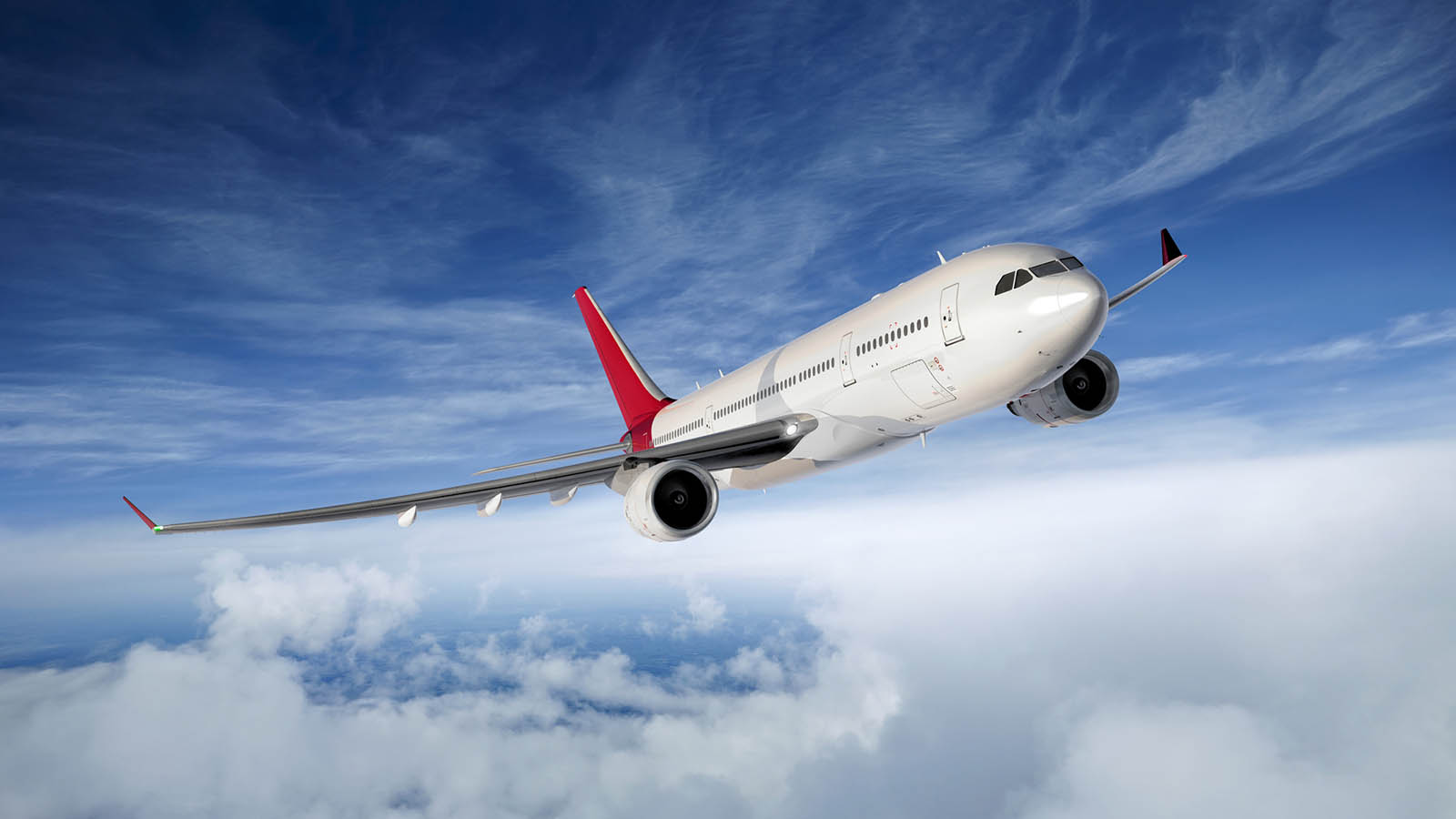}
    \caption{}
  \end{subfigure}
  \begin{subfigure}[T]{0.19\linewidth}
    \includegraphics[width=\linewidth, height=0.6\linewidth]{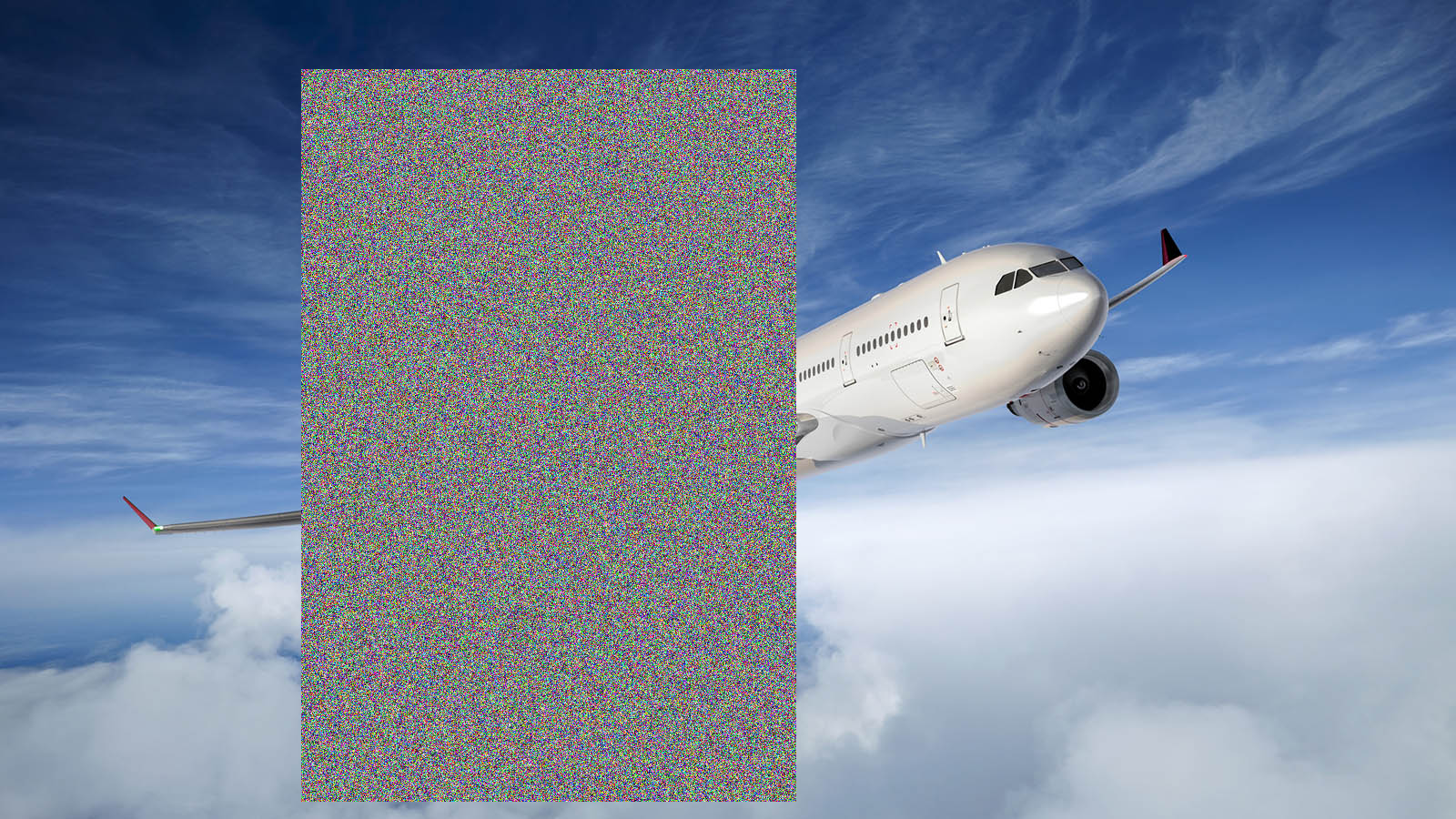}
    \caption{}
  \end{subfigure}
  \begin{subfigure}[T]{0.19\linewidth}
    \includegraphics[width=\linewidth, height=0.6\linewidth]{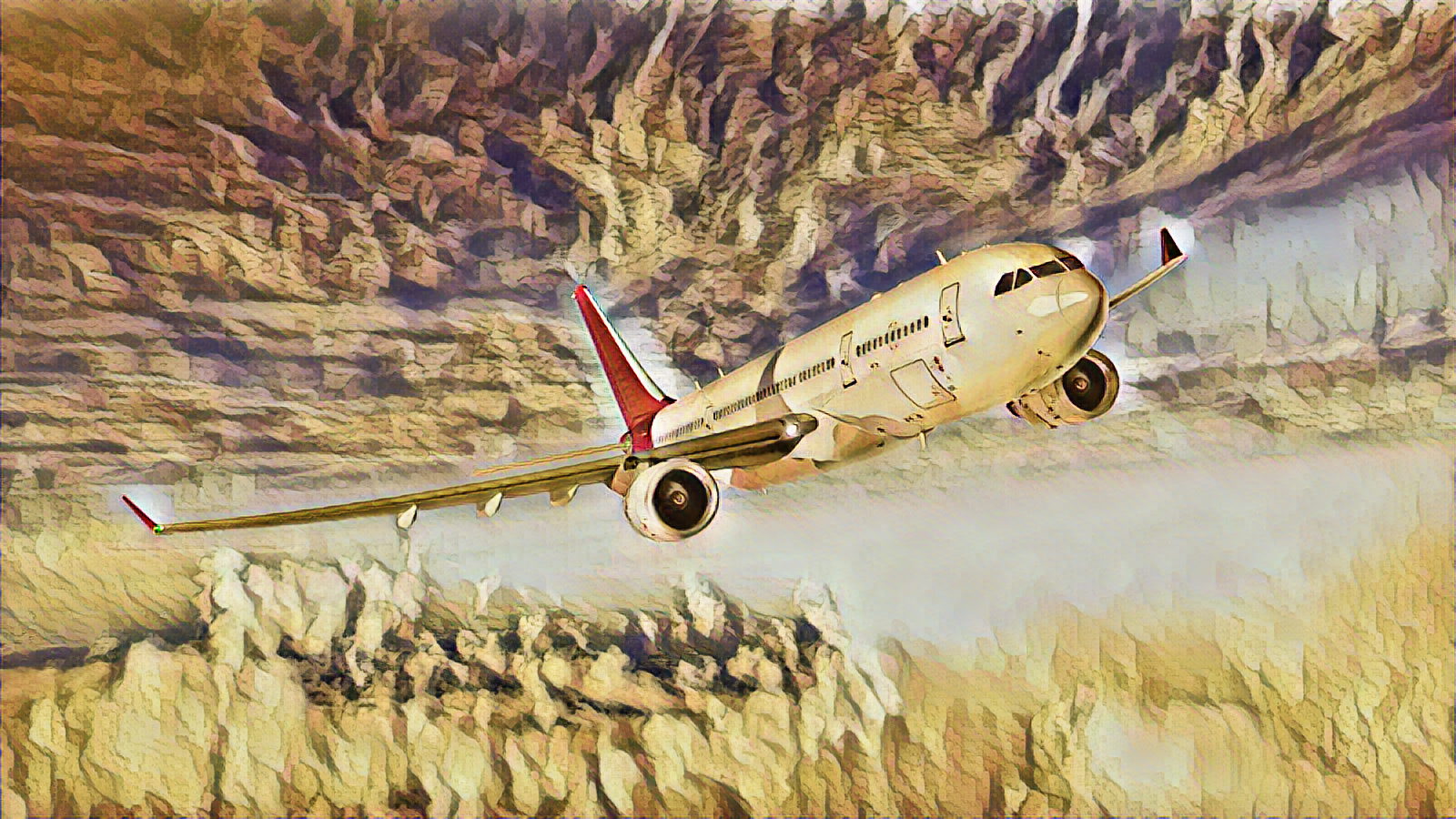}
    \caption{}
  \end{subfigure}
  \begin{subfigure}[T]{0.19\linewidth}
    \includegraphics[width=\linewidth, height=0.6\linewidth]{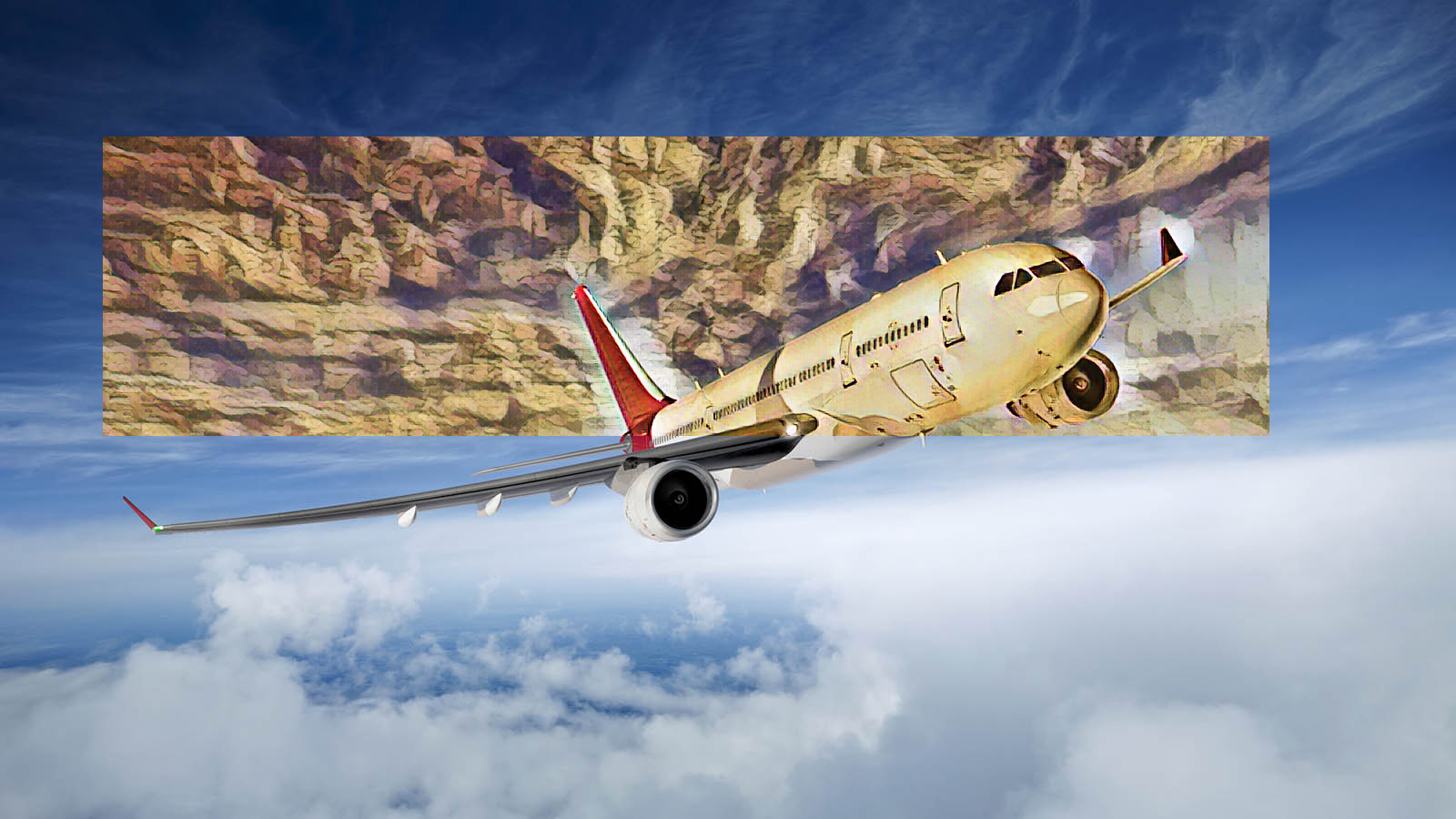}
    \caption{}
  \end{subfigure}
  \begin{subfigure}[T]{0.19\linewidth}
    \includegraphics[width=\linewidth, height=0.6\linewidth]{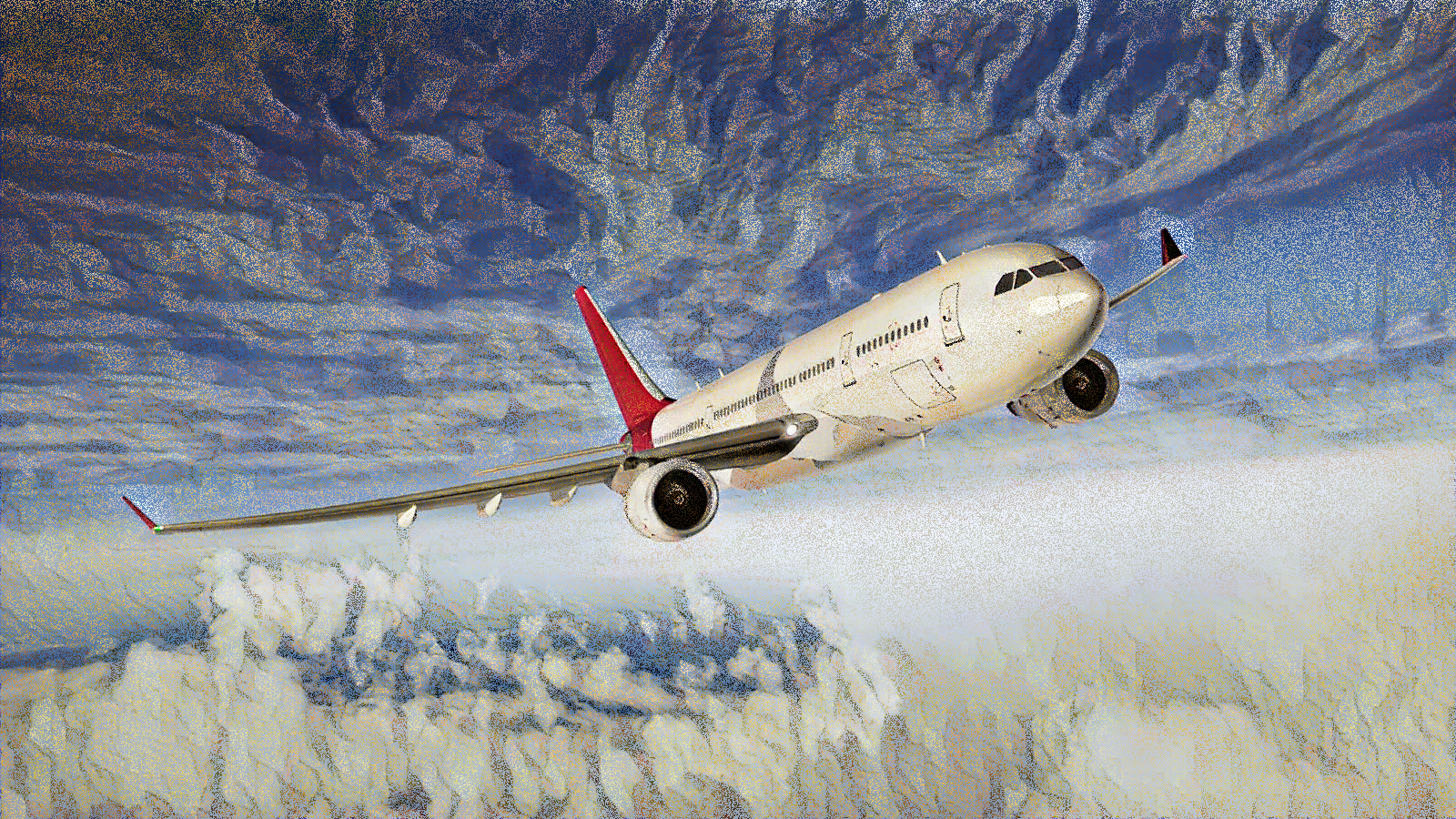}
    \caption{}
  \end{subfigure}
  \caption{Examples of random style transfer: we generate a style-transferred image and use it to patch original image in different ways (a) Input Image. (b) Random-Region-Erased Image. (c) Style Transfer. (d) Random Style Replacement. (e) Random Style Replacement.}
  \label{fig:cmp-study}
\end{figure}

Data augmentation~\cite{yang2024comparative} has garnered significant attention in supervised learning research across a wide range of domains—including computer vision \cite{ding2022data, kumar2024image}, natural language processing \cite{Bayer2022}, graph learning \cite{ sankar2022self, narang2021ranking}, and large language model \cite{jin2021representation, li2024focused} — due to its ability to increase both the volume and diversity of training data, thereby enhancing model generalization and mitigating overfitting. Broadly, data augmentation strategies can be grouped into two categories: generative methods, which utilize models like Variational Autoencoders (VAEs)~\cite{kingma2013auto}, Generative Adversarial Networks (GANs)~\cite{goodfellow2014gan}, Large Language Models (LLMs), or diffusion-based frameworks~\cite{ho2020denoising} to synthesize new data; and traditional methods, which rely on transformations such as random cropping, flipping, rotations, color jittering, and histogram equalization to modify existing samples. While both approaches aim to expose the model to a wider variety of conditions, thus reducing overfitting, traditional augmentation strategies may not fully capture the complexity of real-world variability. Consequently, several studies have explored more refined methods, including style augmentation and random erasing \cite{jackson2019style, zhong2020random, xu2024style}. Style augmentation employs style transfer~\cite{ding2024regional} to alter the visual attributes of training images while preserving their semantic content, thereby increasing robustness to differences in texture, color, and contrast. Random erasing, on the other hand, randomly occludes or replaces subregions of an image, making models more resilient to missing or corrupted information. In this paper, we revisit these traditional approaches—particularly focusing on their potential to advance the efficacy of data augmentation in supervised learning.

In this paper, we introduce a novel data augmentation method that merges style augmentation with random erasing. Our approach involves applying a random style transfer to an image, followed by replacing specific subregions with patches from the style-transferred version. This technique enhances robustness against style variations and occlusion-like effects. It integrates smoothly with existing style transfer frameworks and fits easily into standard data augmentation pipelines, as illustrated in Figure \ref{fig:cmp-study}.

The main contributions of our work are as below:
\begin{enumerate}
    \item We propose a technique that merges style augmentation and random erasing, offering benefits from both texture variation and structured occlusion.
    \item  We demonstrate through experiments that our approach reduces the risk of overfitting while achieving faster convergence compared to established style augmentation methods.
Ease of Integration: Our strategy is parameter-free and can be readily adapted to a broad spectrum of computer vision tasks, making it a highly practical solution for data augmentation.
\end{enumerate}

By leveraging this new augmentation method, we observe notable gains in model performance across different tasks, highlighting its potential to address the persistent challenge of limited labeled data in computer vision research.

\section{Dataset}
We tested our random style replacement method on the STL-10 dataset, which includes 5,000 training images and 8,000 test images, each with a resolution of 96×96 pixels across 10 classes \cite{coates2011analysis}. We chose STL-10 due to its complex backgrounds and high resolution, which pose a substantial challenge for image classification, making it a robust benchmark. Additionally, the limited size of the training set highlights the effectiveness of our data augmentation technique in enhancing training data.

\section{Methods}
This sections introduces our random style replacement method in details. We described the overall process of random style replacement and explain how we perform image patch.
\subsection{Random Style Replacement}
During training, random style replacement is applied with a certain probability $p$: for each image $I$ in a mini-batch, there's a probability $p$ that it undergoes style replacement and a
probability $1-p$ that it remains unchanged.

If selected, the image will be transformed into a new version with a partial style change. This random style replacement process consists of two steps: generating a complete-style-transferred image and merging it with the original image by certain patching methods. The procedure is shown in Alg. \ref{algorithm 1}.

Style transfer refers to a class of image processing algorithms that alter the visual style of an image while preserving its semantic content. For style transfer to be part of a data augmentation technique, it needs to be a both fast and random algorithm capable of applying a broad range of styles. Therefore, we adopt the approach of Jackson et al., which efficiently generates a completely style-transferred image by incorporating randomness on the fly without requiring heavy computations \cite{jackson2019style}.

The generated style-transferred image will then be used to patch the original image, creating an augmented image. There are multiple patching methods, and we adopt the two most common ones: patching by a random subregion and patching randomly selecting individual pixels. To avoid bias in data augmentation, we employed random style transferring to ensure diverse and uniform modifications across all image types, enhancing model generalization.

\begin{algorithm}[t]
\SetAlgoLined
\SetKwInOut{Input}{Input}
\SetKwInOut{Output}{Output}
\SetKwInput{Initialization}{Initialization}
\caption{Random Style Replacement Procedure}\label{algorithm 1}
\Input{Input image $I$; \\
       Augmentation probability $p$; \\
       Patch mode $pMode$;
       }
\Output{Augmented image $I^{\ast}$.}
\Initialization{$p_1 \leftarrow $ Rand (0, 1).} 

\eIf{$p_1 \geq p$}{
   $I^{\ast} \leftarrow I$; \\
   \Return{$I^{\ast}$}.
}{
   $I^{'} \leftarrow \textbf{randomStyleTransfer}(I)$; \\
   $I^{\ast} \leftarrow \textbf{randomPatch}(I, I^{'}, pMode)$; \\
   \Return{$I^{\ast}$}.
}
\end{algorithm}

\subsection{Random Patch}
Random patch is to patch a image based on another image. Here, we provided a detailed explanation of random patch by subregion. This method copies a randomly selected region from the style-transferred image onto the original image. Specifically, it randomly selects a rectangle region $I_e$ within the image and overwrite all its pixels with those from the style-transferred image. 

Firstly we will determine the shape of the patching area $I_e$. Assume the training image has dimensions $W \times H$ and an area $S=W \times H$. We randomly initialize the area of the patched rectangle region to $S_e$, where $\frac{S_e}{S}$ falls within the range defined by the minimum $s_l$ and maximum $s_h$. Similarly, the aspect ratio of the rectangle region, denoted as $r_e$, is randomly chosen between $r_l$ and $r_h$. Given those, the dimensions of $I_e$ are computed as $H_e = \sqrt{S_e \times r_e}$ and $W_e = \sqrt{\frac{S_e}{r_e}}$. 

Next, we randomly select a point $\mathcal{P} = (x_e, y_e)$ within $I$ to serve as the lower-left corner of $I_e$. If the selected region $I_e$ are completely inside $I$ (i.e. $x_e + W_e \le W$ and $y_e + H_e \le H$), we define it as the selected rectangular region. Otherwise, we repeat the selection process until a valid $I_e$ is found. The whole procedure for selecting the rectangular region and applying the patch to original image is illustrated in Alg. \ref{algorithm 2}.
\begin{algorithm}[t]
\SetAlgoLined
\SetKwInOut{Input}{Input}
\SetKwInOut{Output}{Output}
\SetKwInput{Initialization}{Initialization}
\caption{Random Patch by Subregion}\label{algorithm 2}
\Input{Input image $I$; \\
       Utility image $I^{'}$; \\
       Patched area ratio range $s_l$ and $s_h$; \\      
       Patched aspect ratio range $r_l$ and $r_h$.}
\Output{Patched image $I^{\ast}$.}
$S_e\leftarrow $ Rand $(s_l, s_h)$$\times S$;\\
$r_e \leftarrow $ Rand $(r_l, r_h)$;\\
 $H_e \leftarrow \sqrt{S_e \times r_e}$,~ $W_e \leftarrow \sqrt{\frac{S_e}{r_e}}$;\\
\While {True}{
      $x_e \leftarrow $ Rand $(0, W)$,~ $y_e \leftarrow $ Rand $(0, H)$;\\
      \If{$x_e + W_e \le W$ and $y_e + H_e \le H$}{
          $I_e \leftarrow (x_e, y_e, x_e+W_e, y_e+H_e)$;\\
          $I(I_e) \leftarrow $ $I^{'}(I_e)$;\\
          $I^{\ast} \leftarrow I$;\\
          \Return{$I^{\ast}$}.
          }
     }
\end{algorithm}

\section{Experiment}
\subsection{Experiment Settings}
As mentioned in previous sections, we evaluated our random style replacement method for image classification using the well-known STL-10 dataset \cite{coates2011analysis}. To ensure the effectiveness and fairness of our evaluation, we set our experiment conditions mostly the same as \cite{jackson2019style, zhong2020random}. The benchmark networks we selected are ResNet18, ResNet50, ResNet101 and ResNet152 without pre-trained parameters \cite{he2016deep}. 

In all experiments, instead of introducing more advanced optimizers or training procedures such as \cite{xu2024stochastic}, we selected the Adam optimizer (momentum $\beta_1$ = 0.5, $\beta_2$ = 0.999, initial learning rate of 0.001) to align with the settings of other data augmentation methods \cite{jackson2019style, zhong2020random, kingma2014adam}. For our setting of the style augmentation parameter, we selected the style interpolation parameter $\alpha$ as 0.5 and augmentation ratio as 1:1 \cite{jackson2019style}. All experiments are trained on RTX 4090 with 100 epochs. 
\begin{enumerate}
    \item Original dataset without any advanced data augmentation techniques.
    \item Dataset with naive data augmentation by simply copying and stacking the original dataset to match the same augmentation ratio as other groups, along with some routine augmentation operations.
    \item Dataset with random style replacement by subregion. 
    \item Dataset with random style replacement at the pixel level (with an independent probability p = 0.5).
\end{enumerate}

\subsection{Classification Evaluation}
We evaluated our proposed data augmentation technique on the STL-10 dataset, with only 5,000 training images and 8,000 test color images. After the augmentation process, as mentioned in previous sections, the size of the augmented training set will double to 10,000 and the corresponding augmentation treatment will be randomly applied accordingly. We applied the same training settings as prior work, whose effectiveness in strategy and hyperparameter selection, including learning rate, has already been verified.

\begin{figure}[h!]
  \centering
    \includegraphics[width=0.9\linewidth, height=0.45\linewidth]{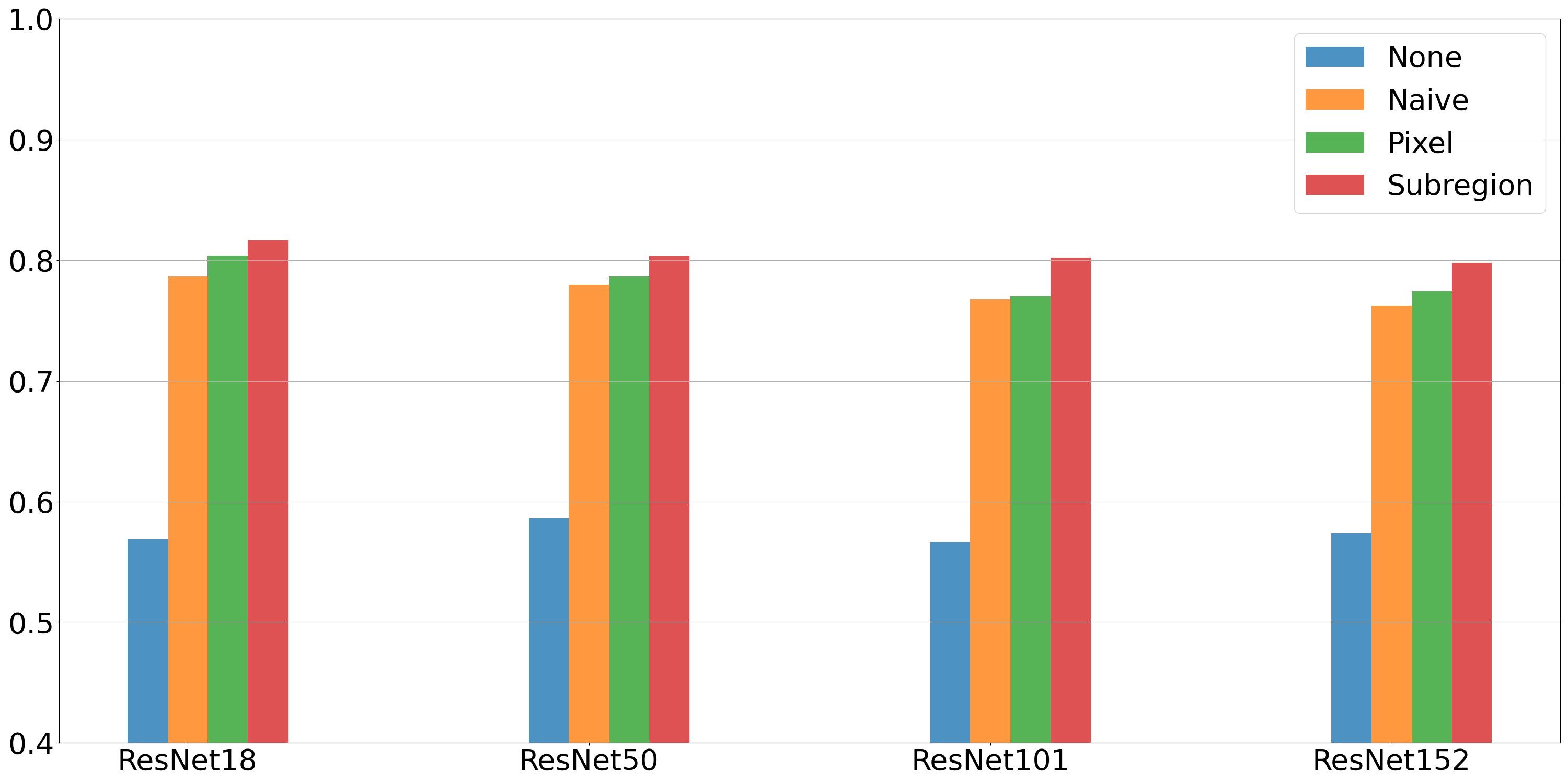}
    \caption{Classification Accuracy of ResNets on STL-10 test set. "None" represents original dataset. "Naive" represents dataset with naive data augmentation by simply stacking the original dataset. "Pixel" represents dataset with random style replacement at the pixel level. "Subregion" represents dataset with random style replacement by subregion.}
  \label{fig:bar}
\end{figure}

Our method achieved 81.6\% classification accuracy in just 100 training epochs, as shown in Fig. \ref{fig:bar}. This result is both faster and more accurate than the 80.8\% accuracy after 100,000 epochs reported by Jackson et al. \cite{jackson2019style}, highlighting our approach's efficiency and scalability.

We also tested our data augmentation technique across various network architectures including ResNet18, ResNet50, ResNet101, and ResNet152, where it consistently outperformed others, demonstrating its robustness and versatility for a wide range of computer vision tasks.

Furthermore, our findings support those of Zhong et al. \cite{zhong2020random}, who found that erasing entire subregions is more effective than pixel-level erasing. Similarly, our data shows that random style replacement within subregions is a superior augmentation strategy, enhancing the training data's representational richness and contributing to faster model convergence and improved performance. This strategy maintains structural integrity and introduces variations that reflect the natural diversity of real-world datasets.

\begin{figure}[h]
\centering
  \begin{subfigure}[T]{0.49\linewidth}
    \includegraphics[width=\linewidth, height=0.5\linewidth]{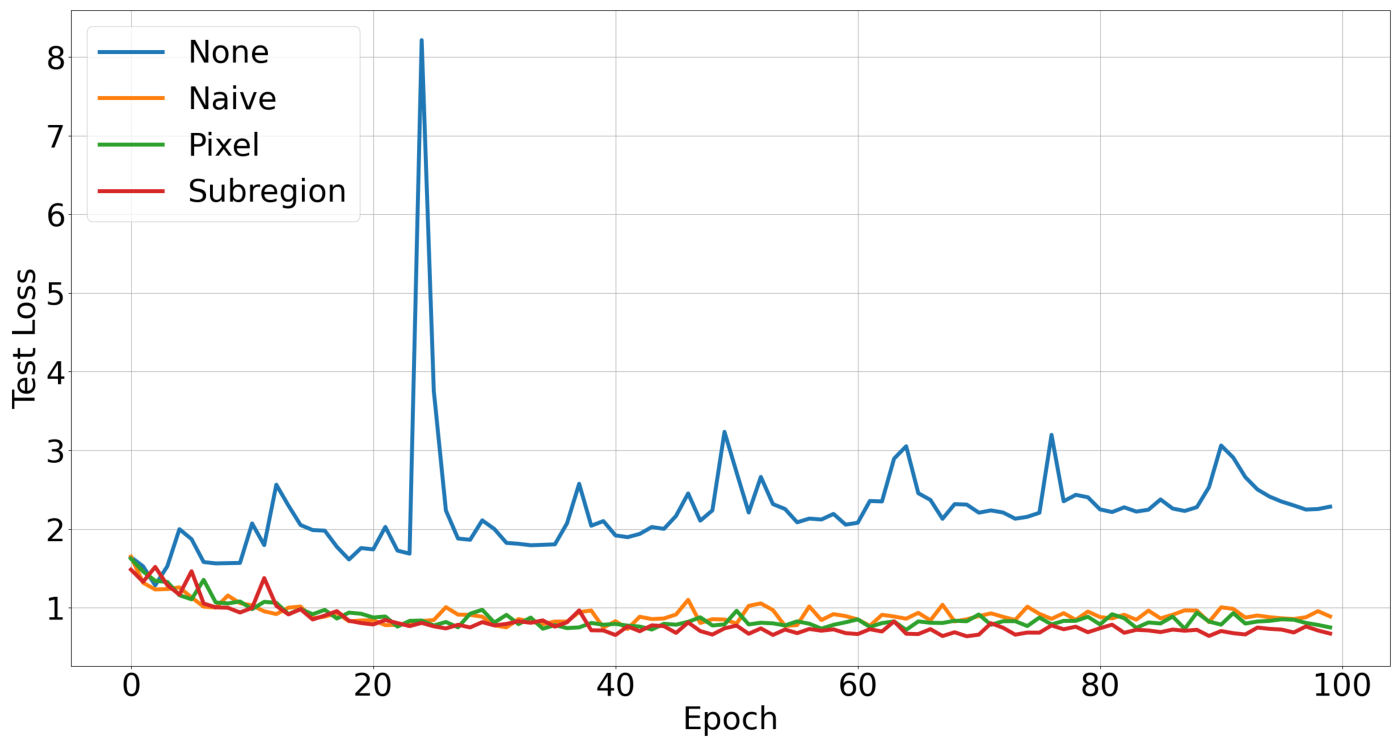}
  \end{subfigure}
  \begin{subfigure}[T]{0.49\linewidth}
    \includegraphics[width=\linewidth, height=0.5\linewidth]{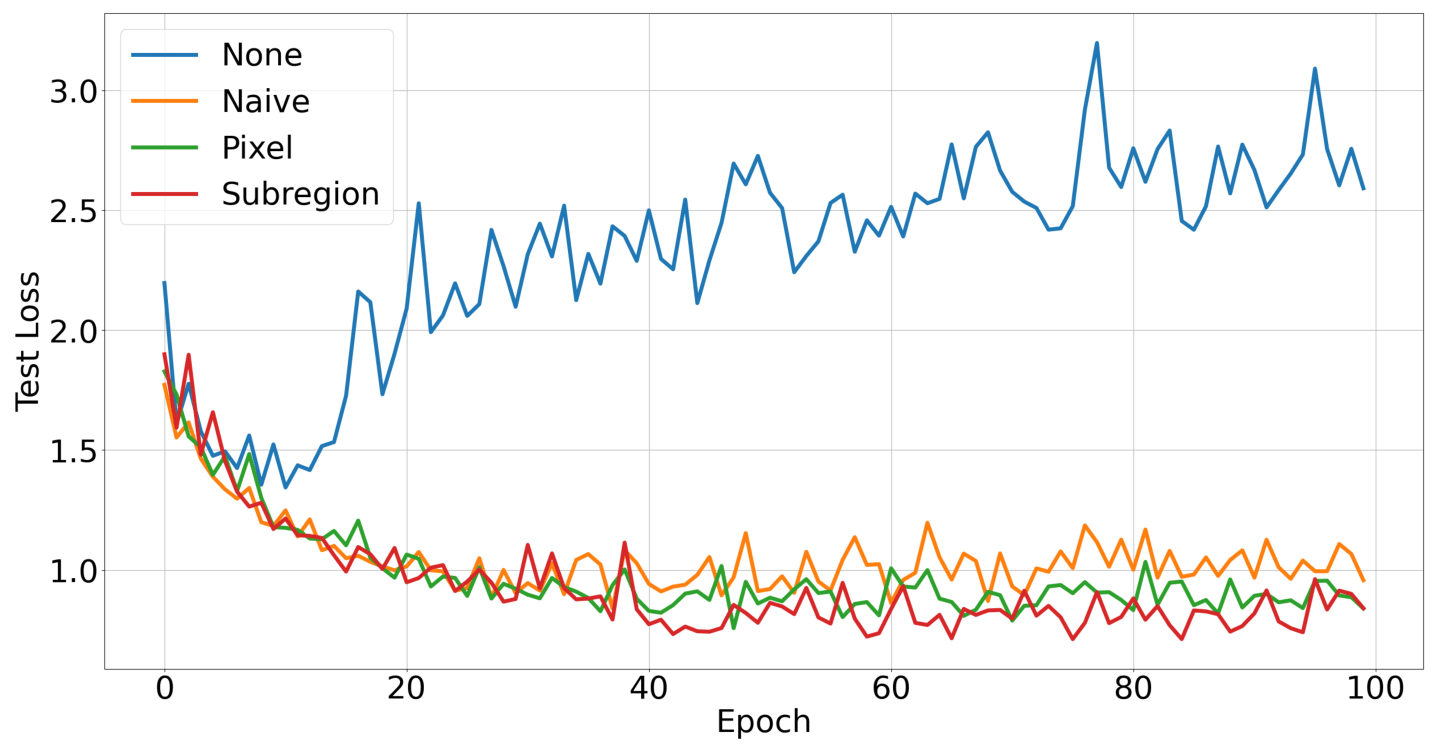}
  \end{subfigure}

  \begin{subfigure}[T]{0.49\linewidth}
    \includegraphics[width=\linewidth, height=0.5\linewidth]{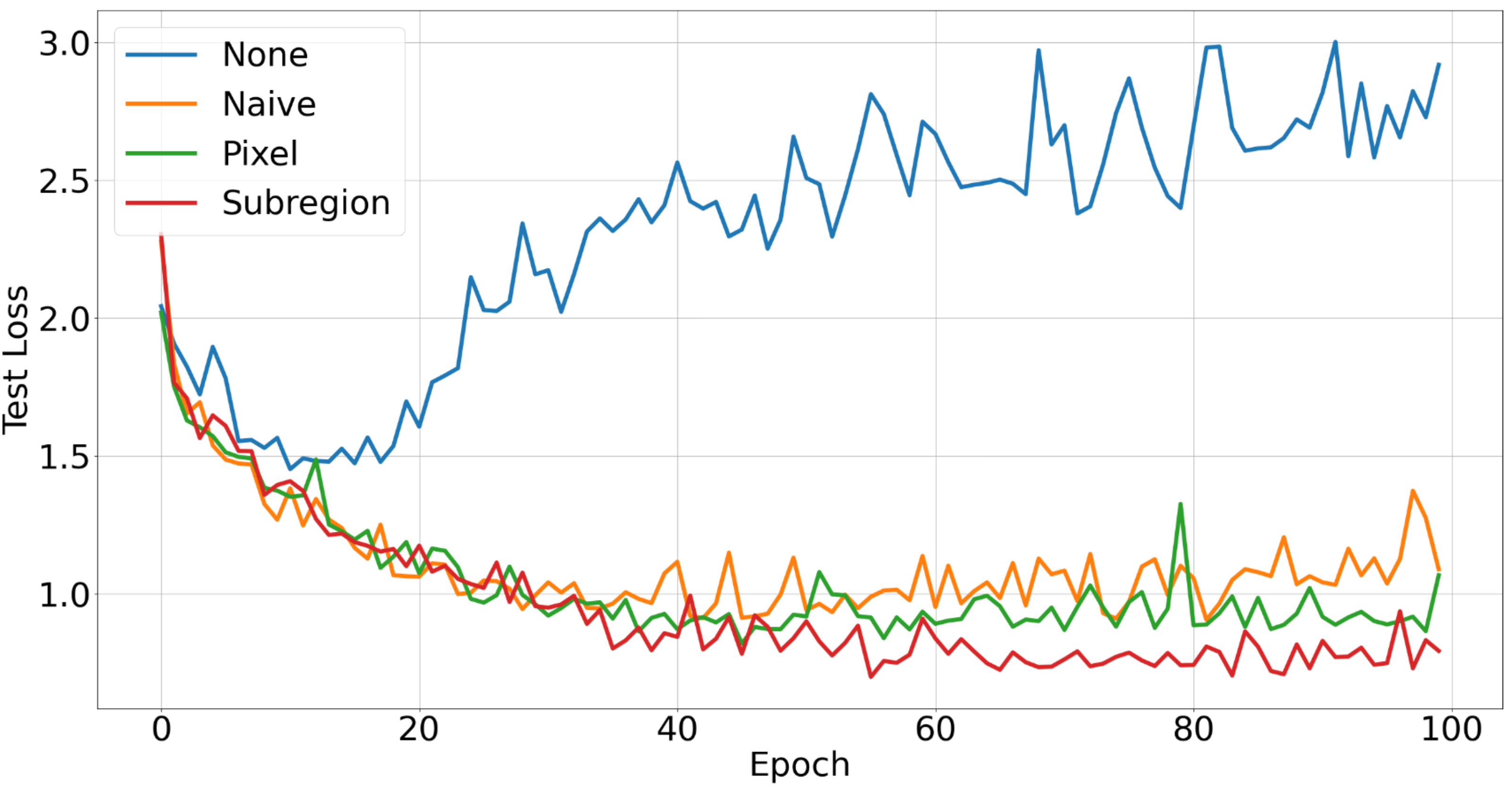}
  \end{subfigure}
  \begin{subfigure}[T]{0.49\linewidth}
    \includegraphics[width=\linewidth, height=0.5\linewidth]{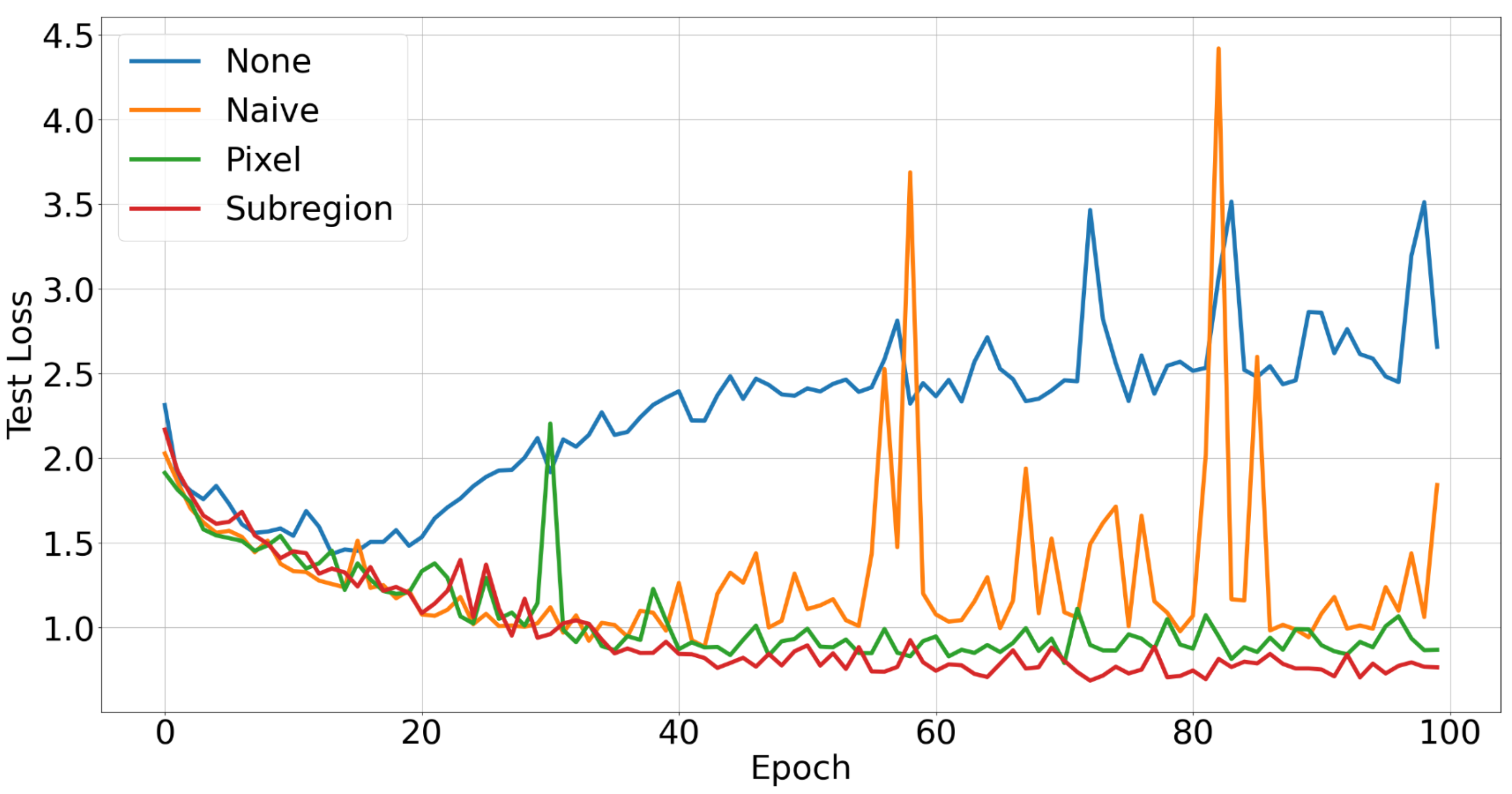}
  \end{subfigure}
    \caption{Loss of ResNets on STL-10 test set. "None" represents original dataset. "Naive" represents dataset with naive data augmentation by simply stacking the original dataset. "Pixel" represents dataset with random style replacement at the pixel level. "Subregion" represents dataset with random style replacement by subregion.}
  \label{fig:loss}
\end{figure}

To confirm the effectiveness of our data augmentation technique, we analyzed the test loss of each method on the STL-10 test set, as shown in Fig. \ref{fig:loss}. In contrast to the naive dataset, whose test loss stops converging after the 20th epoch, all augmented datasets show improved convergence speed and reduced loss variability. Notably, the style augmentation strategy that randomly replaces subregions achieves the fastest convergence and the most stable training process. Despite varying effectiveness in stabilizing training loss across ResNets, our method's performance remains consistently stable.

\section{Conclusions}
In conclusion, our random style replacement strategy offers a practical and scalable data augmentation solution for the STL-10 dataset and beyond. By innovatively combining \cite{jackson2019style} and \cite{zhong2020random}, our proposed data augmentation framework demonstrates superior performance and achieves faster convergence. By randomly replacing subregions rather than individual pixels, we preserve critical structural information while introducing meaningful variability, resulting in faster training convergence and higher accuracy. Our experiments with multiple ResNet architectures consistently verify the robustness of this method, showcasing its versatility for diverse computer vision applications.

\section{Future Work}
Our random style replacement method has shown promising results, yet further validation is needed to confirm its wider applicability. It is crucial to test this technique across various datasets and tasks to establish its generalizability and identify any limitations. Additionally, the convergence speed observations require confirmation through further experiments involving diverse datasets and network architectures. Moreover, integrating Large Language Models (LLMs) guided approaches \cite{ding2024enhance} could enhance the method. These approaches would use LLMs to guide style replacement, potentially selecting optimal subregions for style transfer based on foreground and background information, thus enabling more meaningful and effective transformations.

\renewcommand{\bibfont}{\footnotesize}

\footnotesize{
\bibliographystyle{IEEEtran}
\bibliography{main}
}

\end{document}